\documentclass[
  reprint,           
  amsmath,amssymb,
  superscriptaddress 
]{revtex4-2}

\usepackage[T1]{fontenc}
\usepackage[utf8]{inputenc}
\usepackage{microtype}
\usepackage{graphicx}
\usepackage{booktabs}
\usepackage{dcolumn}
\usepackage{bm}
\usepackage{siunitx}
\usepackage{hyperref}
\usepackage{cleveref}
\usepackage{inconsolata}

\begin{document}

\title{Founder effects shape the evolutionary dynamics of multimodality in open LLM families}

\author{Manuel Cebrian}
\affiliation{Center for Automation and Robotics, Spanish National Research Council, Madrid, Spain}

\begin{abstract}
Large language model (LLM) families are improving rapidly, yet it remains unclear how quickly multimodal capabilities emerge and propagate within open families. Using the ModelBiome AI Ecosystem dataset of Hugging Face model metadata and recorded lineage fields ($>1.8\times 10^{6}$ model entries), we quantify multimodality over time and along recorded parent-to-child relations. Cross-modal tasks are widespread in the broader ecosystem well before they become common within major open LLM families: within these families, multimodality remains rare through 2023 and most of 2024, then increases sharply in 2024--2025 and is dominated by image--text vision--language tasks. Across major families, the first vision--language model (VLM) variants typically appear months after the first text-generation releases, with lags ranging from $\sim$1 month (Gemma) to more than a year for several families and $\sim$26 months for GLM. Lineage-conditioned transition rates show weak cross-type transfer: among fine-tuning edges from text-generation parents, only 0.218\% yield VLM descendants. Instead, multimodality expands primarily within existing VLM lineages: 94.5\% of VLM-child fine-tuning edges originate from VLM parents, versus 4.7\% from text-generation parents. At the model level, most VLM releases appear as new roots without recorded parents ($\sim$60\%), while the remainder are predominantly VLM-derived; founder concentration analyses indicate rapid within-lineage amplification followed by diversification. Together, these results show that multimodality enters open LLM families through rare founder events and then expands rapidly within their descendant lineages, producing punctuated adoption dynamics that likely induce distinct, transfer-limited scaling behavior for multimodal capabilities.
\end{abstract}

\keywords{multimodality; vision--language models; model ecosystems; lineage analysis; diffusion} 

\maketitle


\section{Introduction}

Progress in foundation models has been propelled by regular performance gains from scale---in parameters, data, and compute---and by post-training methods that make general pretrained models broadly usable \cite{Bommasani2021FoundationModels,Brown2020GPT3,Kaplan2020ScalingLaws,Hoffmann2022Chinchilla,Wei2021FLAN,Ouyang2022InstructGPT}. In parallel, an open ``derivative'' ecosystem has formed around model hubs, where base checkpoints are repeatedly adapted through fine-tuning, quantization, and merging. This produces large, time-evolving families of related models, and lowers the cost of reuse and recombination of capabilities across communities \cite{Wolf2020Transformers}.

A central frontier in this ecosystem is multimodality, particularly image--text reasoning. Vision--language models (VLMs) have advanced rapidly by coupling vision encoders to large language models and training or tuning on large-scale image--text corpora \cite{Alayrac2022Flamingo,Chen2022PaLI,Li2023BLIP2,Liu2023LLaVA}. Yet multimodality is not a trivial extension of text-only development: it requires additional data pipelines, architectural interfaces, and evaluation protocols, and it introduces distinct reliability problems in grounding and visual faithfulness (e.g., object hallucination) \cite{Li2023POPE,Zhou2023MitigatingHallucination}. These requirements suggest that the mechanism by which multimodality appears in open model families may differ from the routine derivative dynamics observed for text-only checkpoints. A basic empirical question follows: in a lineage-rich open ecosystem, does multimodality primarily arise via incremental adaptation of text-only checkpoints, or via less frequent integration events that create VLM ``founders'' followed by within-lineage expansion?

\begin{figure*}[!t]
\centering
\includegraphics[width=\linewidth]{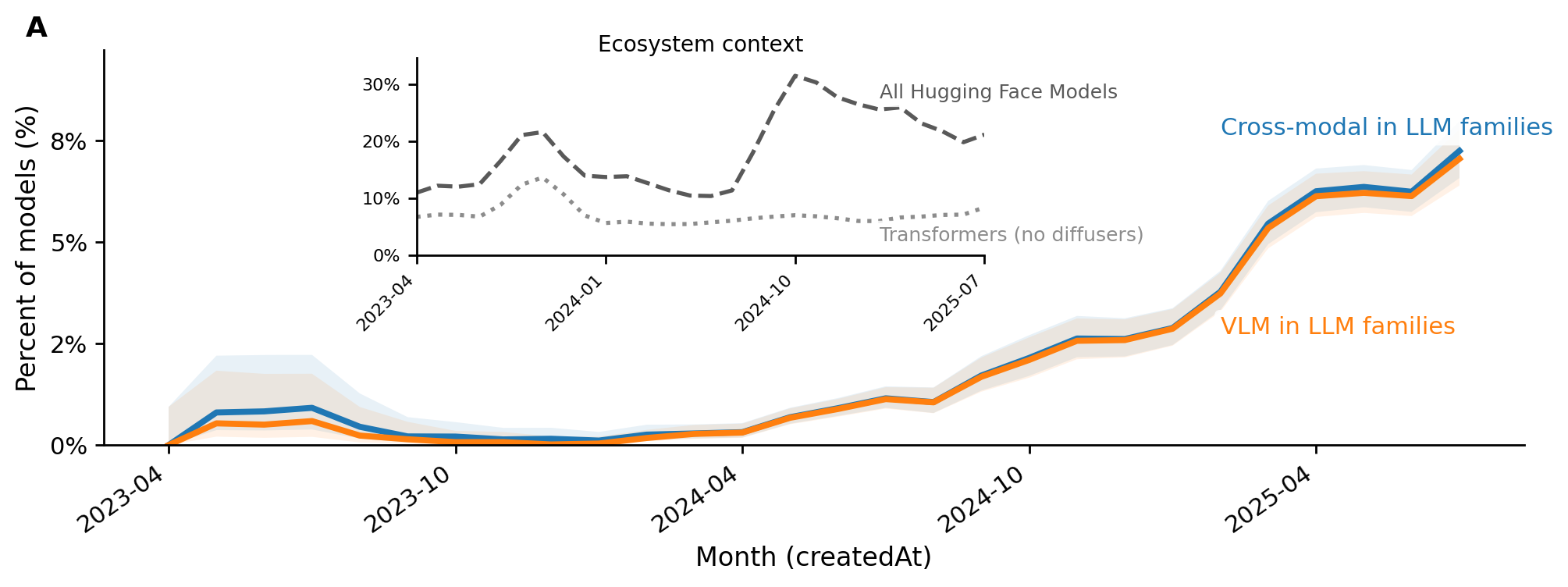}
\caption{\textbf{Multimodality appears earlier in the broader ecosystem than within major open LLM families.}
Main panel: for each month, the share of newly created checkpoints in major open LLM families tagged with any cross-modal task (text paired with image/audio/video; blue) and the share tagged specifically with image--text vision--language tasks (orange). Inset: corresponding ecosystem-wide reference series for all task-tagged Hugging Face models (dashed) and for \texttt{transformers} models excluding diffusion-oriented pipelines (dotted). LLM families are identified by name-based \texttt{model\_id} patterns within \texttt{transformers} (excluding diffusers); months with low volume are omitted (e.g., $n<300$). Shaded bands show 95\% Wilson score confidence intervals.}

\label{fig:multimodality_trends}
\end{figure*}

Addressing this question requires ecosystem-scale measurement of both timing and transmission: (i) when multimodal traits become prevalent within and across model families, and (ii) how these traits propagate along explicit parent--child relations. A key enabling development is the mapping of the Hugging Face model hub as an evolutionary ecosystem with millions of models and recorded relationship fields linking derivatives to their parents \cite{Laufer2025Ecosystem}. This ecosystem is complemented by semi-structured documentation in model cards \cite{Mitchell2019ModelCards}, which, despite noise and heterogeneity, provides population-level signals about intended use and modality.

Here we use the ModelBiome AI Ecosystem dataset---a snapshot of $1.86\times10^{6}$ public Hugging Face models with metadata, model cards, and recorded lineage edges \cite{Laufer2025Ecosystem}---building on recent ecosystem-scale measurement work that has begun to quantify macrodynamics of model production and evaluation infrastructure that were previously difficult to observe at scale \cite{cebrian2025emergent}---to characterize the emergence and diffusion of multimodality in open LLM families.
 First, we characterize ecosystem-level and family-specific temporal trends, showing that cross-modal tasks are common in the broader hub well before they become prevalent within major Transformer-based LLM families. Second, we estimate lineage-conditioned transition rates for the emergence and persistence of VLM traits under different relationship types (fine-tuning, merging, adapters, quantization), revealing weak transfer from text-only parents to VLM descendants and high persistence within VLM lineages. Third, we analyze founder structure in VLM lineages, documenting a large fraction of VLM releases that enter as new roots and a subsequent pattern of rapid within-lineage amplification followed by diversification. Together, these measurements indicate that multimodality in open LLM families is strongly shaped by founder-driven VLM lineages and limited transfer from text-only checkpoints, consistent with punctuated diffusion of multimodal technical innovations through integration events and lineage expansion.

\subsection{Ecosystem-level multimodality precedes adoption in open LLM families}

We quantify the timing of multimodality in major open LLM families relative to the broader Hugging Face ecosystem using the ModelBiome AI Ecosystem dataset, which links model metadata (including task tags and model cards) to recorded parent--child lineage fields. The July 2025 snapshot contains $1.86\times 10^{6}$ model entries and $3.02\times 10^{6}$ directed lineage relations. Because Hugging Face began recording \texttt{createdAt} timestamps on 2 March 2022 and earlier uploads were backfilled to that date, monthly trends are most interpretable from March 2022 onward.

For each calendar month, we compute the fraction of newly created models assigned to (i) cross-modal tasks (text paired with image/audio/video) and (ii) the subset of vision--language tasks (image--text). Proportions are reported with 95\% Wilson score intervals, and months with low model volume are excluded to avoid unstable estimates (e.g., $n<300$).

Figure~\ref{fig:multimodality_trends} shows that multimodal task tags are present at substantial levels in the ecosystem early in the record, both in the full set of task-tagged models and when restricting to \texttt{transformers} models while excluding diffusion-oriented pipelines. In contrast, the same measurement within major open LLM families remains near zero through 2023 and most of 2024 before increasing sharply in 2024--2025.

Within LLM families, the cross-modal series closely tracks the vision--language series, indicating that the observed increase is driven primarily by image--text capability rather than broad uptake of audio/video modalities. This lag between ecosystem-wide multimodality and within-family adoption motivates the lineage-conditioned analyses below, which test whether multimodality enters these families via routine adaptation of text-only checkpoints or via founder-driven VLM lineages.

\begin{figure*}[!t]
\centering
\includegraphics[width=\linewidth]{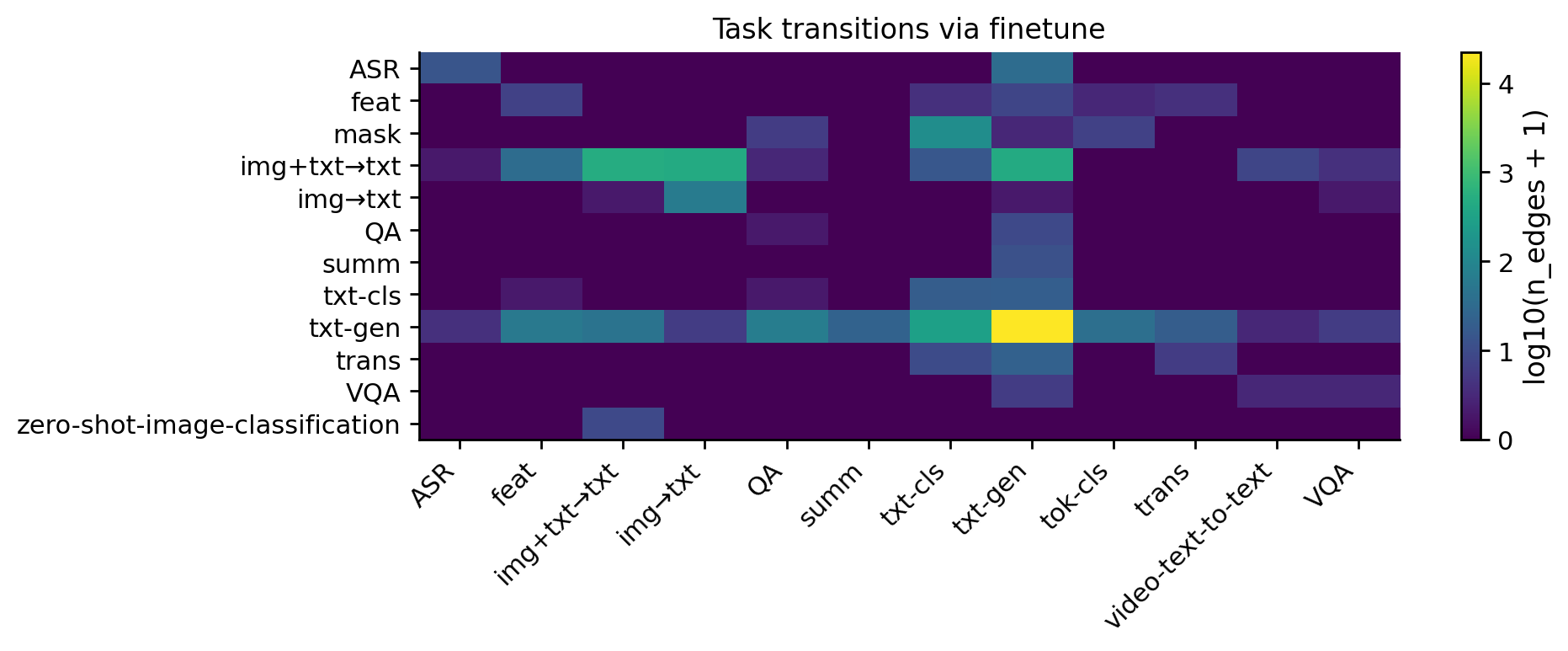}
\caption{\textbf{Task transitions via fine-tuning edges.}
Heatmap of parent$\rightarrow$child task-tag transitions along recorded \texttt{finetune\_parent} relations (log scale; $\log_{10}(n_{\mathrm{edges}}+1)$). Rows correspond to the parent model’s \texttt{pipeline\_tag}, and columns to the child model’s \texttt{pipeline\_tag}. The pronounced diagonal structure indicates that fine-tuning is predominantly task-preserving, with especially strong \texttt{text-generation}$\rightarrow$\texttt{text-generation} continuity. Off-diagonal entries are comparatively sparse, revealing that cross-task transitions are rare. Short axis labels denote Hugging Face \texttt{pipeline\_tag} abbreviations: \texttt{txt-gen} (text-generation), \texttt{txt-cls} (text-classification), \texttt{tok-cls} (token-classification), \texttt{ASR} (automatic-speech-recognition), \texttt{QA} (question-answering), \texttt{summ} (summarization), \texttt{trans} (translation), \texttt{feat} (feature-extraction), \texttt{mask} (fill-mask), \texttt{img$\rightarrow$txt} (image-to-text), \texttt{img+txt$\rightarrow$txt} (image-text-to-text), and \texttt{VQA} (visual-question-answering).}
\label{fig:task_transitions_finetune}
\end{figure*}

\subsection{Lineage transitions reveal weak transfer from text-only checkpoints to VLMs}

We next test whether the late rise of VLMs within open LLM families (Fig.~\ref{fig:multimodality_trends}) can be accounted for by routine lineage transitions from text-only checkpoints. Using recorded parent--child relations, we examine edges in which the \emph{parent} is tagged as \texttt{text-generation} and the \emph{child} is tagged with an image--text vision--language task. We treat relation types---fine-tuning, merging, adapters, and quantization---as distinct channels through which task capabilities may propagate.

Figure~\ref{fig:task_transitions_finetune} summarizes task transitions among fine-tuning edges. The transition mass is strongly task-preserving: most edges fall on the diagonal, dominated by \texttt{text-generation}$\rightarrow$\texttt{text-generation}, consistent with fine-tuning being used primarily to specialize models within an existing task regime. Cross-task transitions are present but concentrated in a small number of pathways (e.g., \texttt{image-to-text}$\leftrightarrow$\texttt{image-text-to-text}), indicating that when task changes occur they are structured rather than diffuse.

Conditioning on \texttt{text-generation} parents, transitions to VLM-tagged children are rare across channels: $0.218\%$ under fine-tuning (50/22{,}928; 95\% Wilson CI 0.165--0.287), $0.104\%$ under merges (12/11{,}594; 0.059--0.181), and $0.133\%$ under quantization (14/10{,}487; 0.080--0.224). The adapter channel has too few observations for stable inference (1/97; Table~\ref{tab:text_to_vlm_rates}). These values quantify weak per-edge transfer from text-only checkpoints into image--text tasks. Because fine-tuning contributes the largest number of recorded edges overall, it also contributes the largest absolute count of observed text-to-VLM events (50 edges) despite its low conditional probability.

We also ask whether cross-modal emergence from \texttt{text-generation} parents extends beyond image--text. Using the broader set of cross-modal tasks (text with image/audio/video), the fine-tuning rate increases only marginally ($0.236\%$, 54/22{,}928; 0.181--0.307), and 92.6\% of these cross-modal events are image--text VLM tasks (50/54). All observed cross-modal transitions via merges and quantization are image--text (12/12 and 14/14). Thus, when text-parent derivatives do exhibit cross-modality, it is overwhelmingly image--text rather than audio- or video-centered.

Together, these results are inconsistent with a dominant ``gradual conversion'' mechanism in which VLM prevalence within open LLM families is primarily produced by frequent transitions from text-generation checkpoints along routine derivative edges. Such transitions exist, but their conditional probability is very low (Table~\ref{tab:text_to_vlm_rates}), and the fine-tuning transition matrix is dominated by task preservation (Fig.~\ref{fig:task_transitions_finetune}). This motivates a complementary growth mode in which multimodality expands mainly within established VLM lineages---with rare bridge events from text-generation parents---which we quantify next by decomposing VLM growth by parent task category and characterizing founder concentration dynamics.

\begin{table}[t]
\centering
\caption{\textbf{Text-generation to VLM transitions by relation type.}
For edges with a \texttt{text-generation} parent, we report the share whose child is a VLM (image--text) task. Intervals are 95\% Wilson score CIs.}
\label{tab:text_to_vlm_rates}

\vspace{5pt}

\small
\setlength{\tabcolsep}{8pt}      
\renewcommand{\arraystretch}{1.12}
\begin{tabular}{@{}lrrr@{}}
\toprule
Relation & $k$ & $n$ & Rate, \% (95\% CI) \\
\midrule
Fine-tune & 50 & 22{,}928 & 0.218 (0.165--0.287) \\
Merge     & 12 & 11{,}594 & 0.104 (0.059--0.181) \\
Quantize  & 14 & 10{,}487 & 0.133 (0.080--0.224) \\
Adapter$^{\dagger}$ & 1 & 97 & 1.031 (0.182--5.611) \\
\bottomrule
\end{tabular}

\vspace{3pt}
\footnotesize $^{\dagger}$Sparse edges; estimate is unstable.
\end{table}

\subsection{Time-resolved transition rates}
\label{subsec:time_resolved_rates}

\begin{figure*}[!t]
\centering
\begin{minipage}[t]{0.49\linewidth}
\centering
\includegraphics[width=\linewidth]{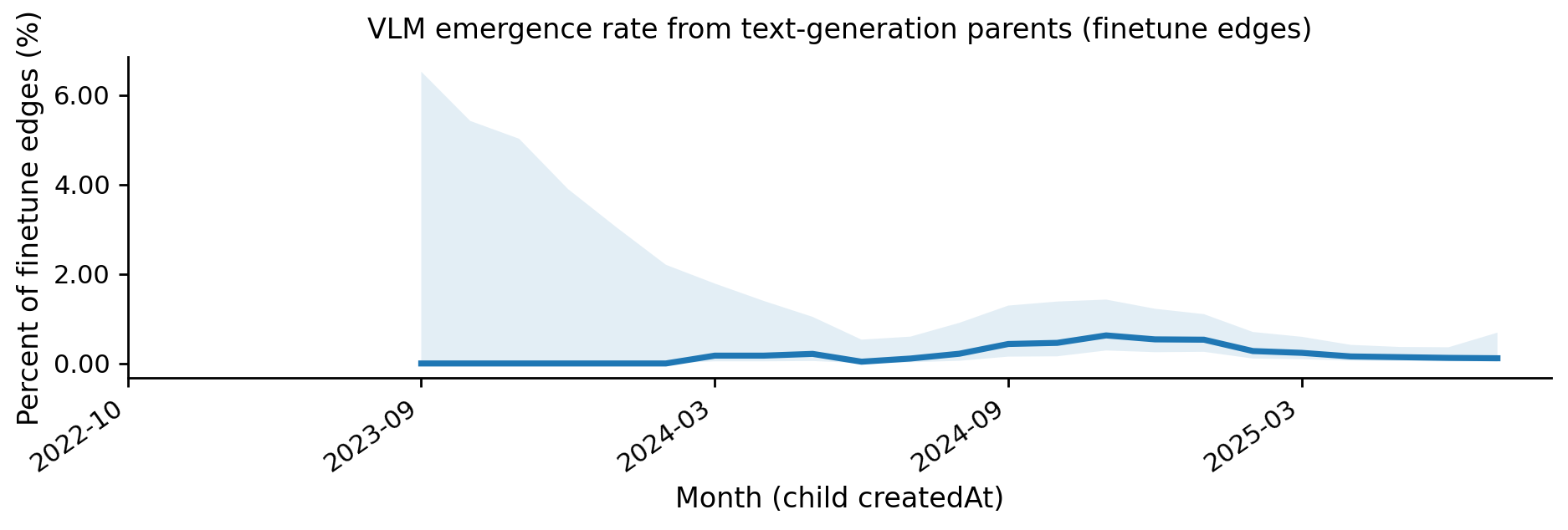}
\end{minipage}\hfill
\begin{minipage}[t]{0.49\linewidth}
\centering
\includegraphics[width=\linewidth]{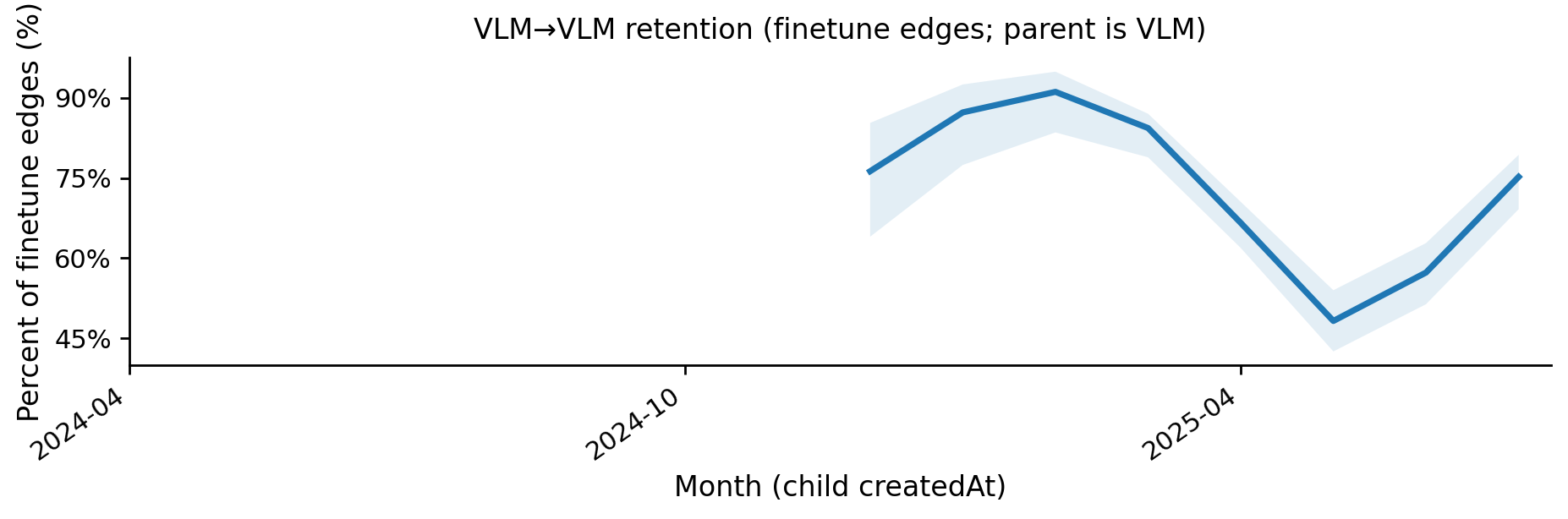}
\end{minipage}
\caption{\textbf{Asymmetric dynamics under fine-tuning: rare text$\rightarrow$VLM emergence but high VLM$\rightarrow$VLM retention.}
Left: Monthly estimates of $P(\mathrm{child\ is\ VLM}\mid \mathrm{parent=text\mbox{-}generation},\,\mathrm{relation=finetune})$, computed over recorded fine-tuning edges and binned by the child model’s \texttt{createdAt} month. Right: Monthly estimates of $P(\mathrm{child\ is\ VLM}\mid \mathrm{parent=VLM},\,\mathrm{relation=finetune})$ (VLM retention), binned by child \texttt{createdAt} month. Shaded bands denote 95\% Wilson score confidence intervals for binomial proportions. Text-to-VLM transitions remain near zero with only transient increases, whereas fine-tuning from VLM parents typically preserves VLM status, indicating strong path dependence in modality along lineage edges.}
\label{fig:finetune_emergence_retention}
\end{figure*}

To test whether the late rise of multimodality within open LLM families (Fig.~\ref{fig:multimodality_trends}) could be driven by an increasing tendency for text-only checkpoints to produce VLM descendants, we estimate a time-resolved, lineage-conditioned transition probability. For each month (indexed by the child's (\texttt{createdAt}), we compute
{\small
\[
P(\mathrm{child\ is\ VLM}\mid \mathrm{parent=text\mbox{-}generation,\,relation=finetune})
\]
}
restricting parents to \texttt{text-generation} models and labeling children as VLM if their pipeline tags correspond to image--text tasks (\texttt{image-to-text} or \texttt{image-text-to-text}). We focus on \texttt{finetune} edges because they dominate recorded lineage relations and most closely represent incremental derivative activity in the open ecosystem. Uncertainty is quantified using 95\% Wilson score intervals.

Figure~\ref{fig:finetune_emergence_retention} summarizes fine-tuning dynamics under two conditioning events. In the left panel, the text$\rightarrow$VLM transition rate remains small throughout the observation window and does not exhibit a sustained upward trend. For much of 2023 and early 2024, monthly estimates are near zero, followed by a brief elevation concentrated in late 2024. The largest spike occurs in November 2024 (10/1{,}061 edges; 0.943\%; 95\% CI 0.513--1.726), with smaller elevations in September 2024 (5/765; 0.654\%; 0.280--1.521) and March 2025 (6/2{,}249; 0.267\%; 0.122--0.581). After this period, the rate returns to near-zero levels in 2025 (e.g., April 2025: 1/4{,}091; 0.024\%; 0.004--0.138; July 2025: 0/339; 0--1.121). Aggregated over the full sample, the overall fine-tuning transition rate is 50/22{,}928 = 0.218\% (95\% CI 0.165--0.287), consistent with the low baseline in the monthly series. By contrast, the right panel shows that conditioning on VLM parents yields substantially higher VLM$\rightarrow$VLM retention over the same period, indicating that once a lineage is multimodal, fine-tuning typically preserves VLM status even though de novo emergence from text-only parents is rare.

These time-resolved estimates rule out a ``gradually increasing conversion'' explanation for the 2024--2025 rise of VLM-tagged models within open LLM families. Instead, direct text-generation-to-VLM conversions appear episodic: rare integration events occur in bursts, but remain infrequent and do not increase monotonically over time. This motivates decomposing the sources of VLM growth by parent task category and lineage structure, to test whether the observed expansion is primarily driven by within-VLM reproduction and founder effects rather than continued conversion from text-generation parents.

\subsection{Founder-driven expansion within VLM lineages}
\label{subsec:founder_vlm}

\begin{figure*}[!t]
\centering

\includegraphics[width=0.48\linewidth]{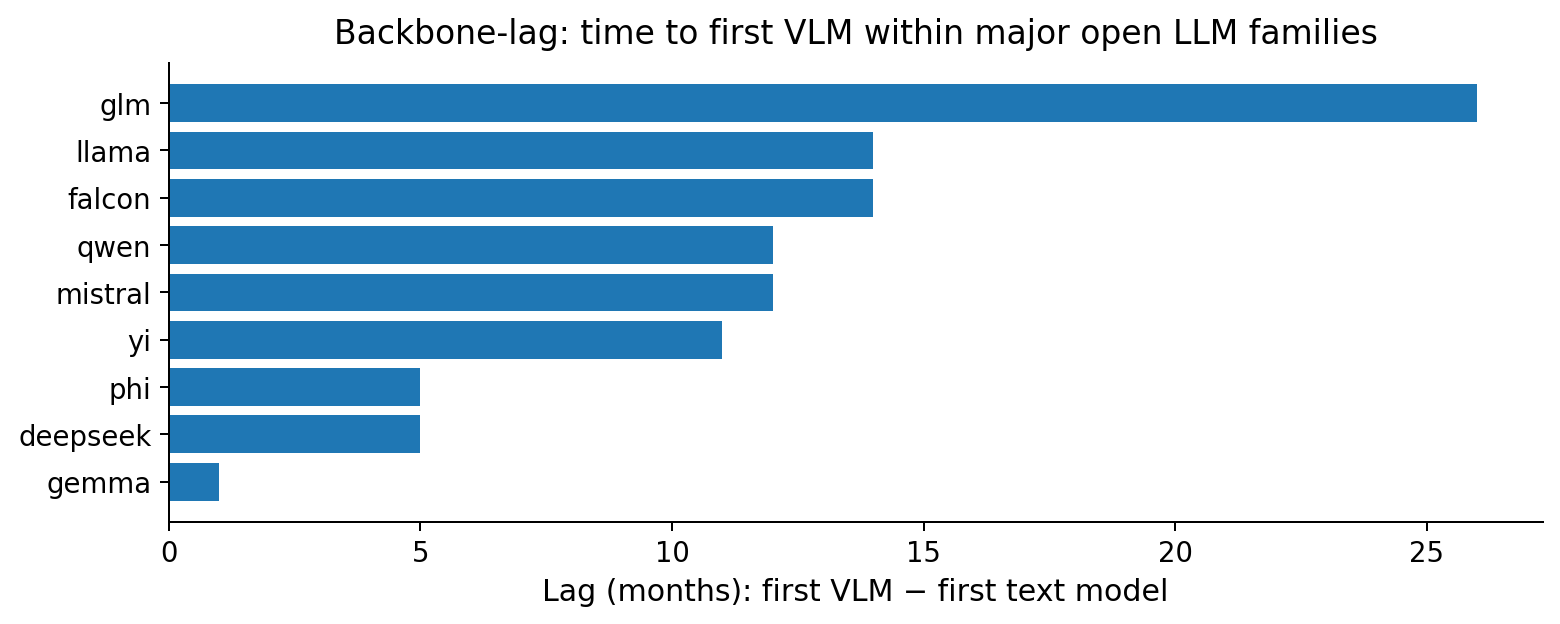}\hspace{0.02\linewidth}%
\includegraphics[width=0.48\linewidth]{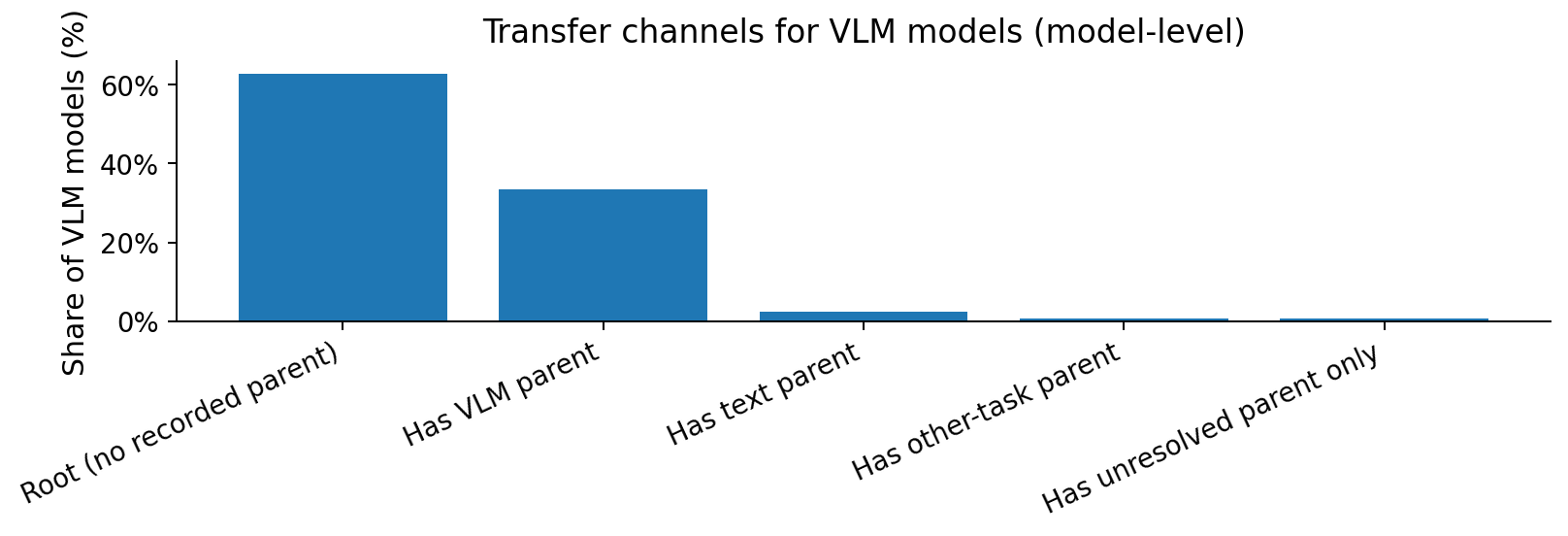}\\[-0.2em]
\includegraphics[width=0.48\linewidth]{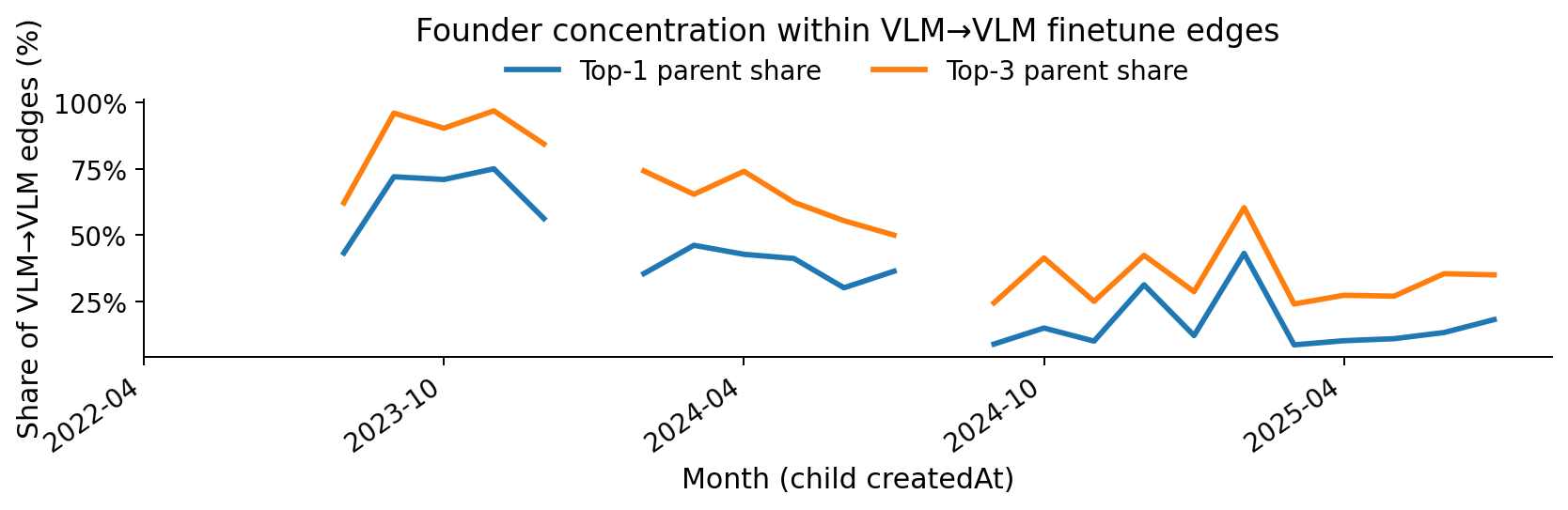}\hspace{0.02\linewidth}%
\includegraphics[width=0.48\linewidth]{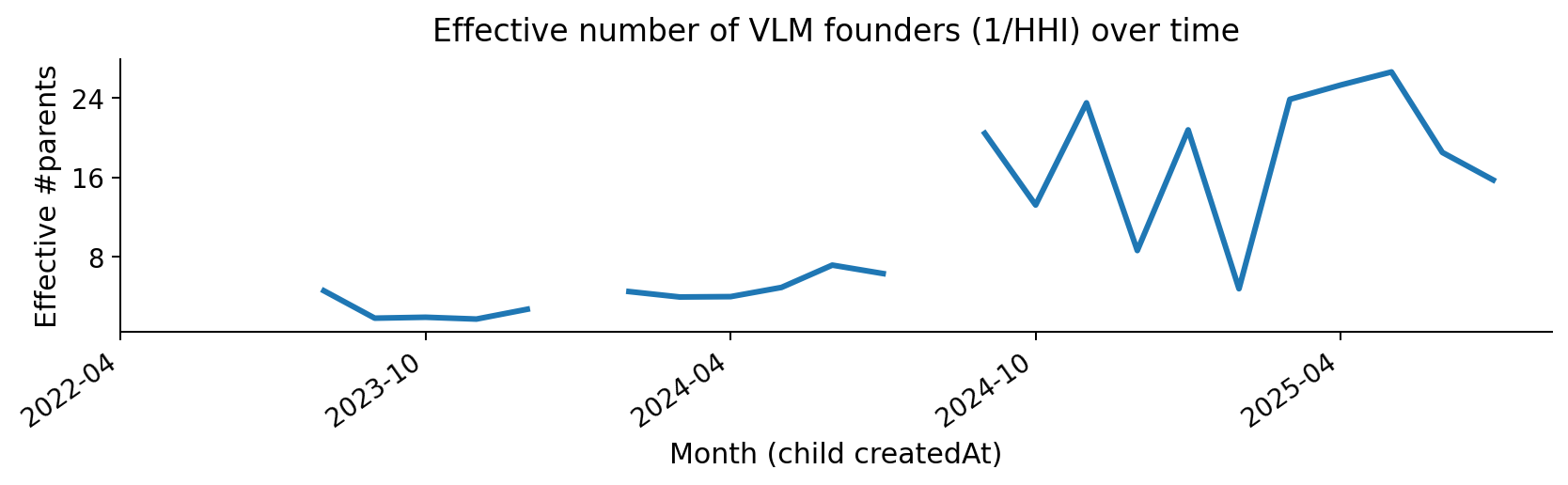}

\caption{\textbf{Founder-driven expansion within VLM lineages.}
(A) Backbone lag to first VLM within major open LLM families, measured as months between the first \texttt{text-generation} release in the family and the first VLM-tagged release.
(B) Model-level lineage channels for VLM releases: ``root'' models have no recorded parent; remaining VLMs are grouped by whether a recorded parent is VLM-, text-, or other-task-tagged (unresolved-parent cases shown separately).
(C) Concentration of VLM$\rightarrow$VLM fine-tuning descent: for each child \texttt{createdAt} month, the share of VLM$\rightarrow$VLM fine-tune edges attributable to the single most prolific parent (top-1) and the three most prolific parents (top-3).
(D) Founder diversity over time, measured as the effective number of parent checkpoints $N_{\mathrm{eff}}=1/\mathrm{HHI}$ computed from the monthly distribution of VLM$\rightarrow$VLM fine-tune parent IDs.
All panels use the ModelBiome AI Ecosystem dataset (July 2025 snapshot).}
\label{fig:founder_vlm}
\end{figure*}

The rapid rise of VLM-tagged models within open LLM families (Fig.~\ref{fig:multimodality_trends}) is not explained by frequent \emph{text}$\rightarrow$\emph{VLM} conversion along recorded lineages. Instead, the recorded lineage structure supports a founder-driven mechanism: once a VLM-capable ancestor exists, VLM labeling is readily preserved and propagated to descendants, while de novo emergence from text-generation parents is exceptionally uncommon.

We quantify this asymmetry by decomposing fine-tuning edges that produce VLM children by the parent task category. Among all fine-tuning edges whose \emph{child} is a VLM, 94.5\% originate from a VLM parent (1005/1063), compared with 4.7\% from text-generation parents (50/1063) and 0.75\% from other-task parents (8/1063). The corresponding conditional rates differ by orders of magnitude: VLM parents frequently yield VLM children (1005/1526; 65.9\%), whereas text-generation parents almost never do (50/22928; 0.218\%). Thus, the VLM growth visible in recorded fine-tuning lineages is dominated by \emph{within-VLM} descent rather than cross-clade transfer from text-only lineages.

Temporal patterns reinforce this interpretation. Conditioning on fine-tuning edges with a VLM parent, the probability that the child remains VLM is typically high (often $\gtrsim 0.75$), but varies over time, consistent with shifts in which founders contribute most edges in a given period (Fig.~\ref{fig:founder_vlm}D). By contrast, the time-resolved \emph{text}$\rightarrow$\emph{VLM} emergence rate remains near zero throughout the window and exhibits, at most, a modest and transient increase in late 2024 rather than a sustained upward trend (Fig.~\ref{fig:founder_vlm}C). Together, these dynamics suggest that the late surge of VLM share within LLM families is better characterized as (i) punctuated introduction of VLM founders, followed by (ii) rapid amplification through VLM$\rightarrow$VLM replication, rather than gradual and widespread conversion of text lineages.

Founder concentration within VLM$\rightarrow$VLM fine-tuning edges is correspondingly strong. A small number of parent checkpoints account for a large fraction of downstream VLM derivatives, consistent with bursty expansion from highly reused founders. As shown in Table~\ref{tab:top_vlm_founders}, the leading parent model (\texttt{naver-clova-ix/donut-base}) accounts for 28.2\% of observed VLM$\rightarrow$VLM edges, and the top three founders together account for 48.9\%. Over time, concentration metrics (top-1 and top-3 parent shares; effective number of founders $N_{\mathrm{eff}}=1/\mathrm{HHI}$) indicate early dominance by a narrow founder set followed by partial diversification, consistent with a classic founder effect in which newly introduced lineages expand rapidly before branching more broadly.

Model-level lineage channels provide an independent, consistent view (Fig.~\ref{fig:founder_vlm}A--B). Most VLM releases appear as roots without recorded parents (approximately 60\%), while the remainder are predominantly derived from existing VLM checkpoints; VLM models with text-only parents constitute a small minority. Taken together, these results support an interpretation of multimodal scaling in open LLM families as bottlenecked by rare cross-clade integration events and dominated, once introduced, by rapid within-VLM lineage expansion.

\begin{table*}[!t]
\centering
\caption{\textbf{Top VLM founders by VLM$\rightarrow$VLM fine-tune descent.}
Parent checkpoints ranked by the number of recorded VLM$\rightarrow$VLM \texttt{finetune\_parent} edges they generate. \emph{Share} is the parent’s fraction of all VLM$\rightarrow$VLM fine-tune edges in this sample, summarizing founder dominance within VLM lineages.}
\label{tab:top_vlm_founders}

\vspace{0.25em} 
\footnotesize
\setlength{\tabcolsep}{7pt} 
\renewcommand{\arraystretch}{1.12} 

\begin{tabular}{r l r r}
\hline\hline
\textbf{Rank} & \textbf{Parent model\_id} & \textbf{Edges} & \textbf{Share (\%)} \\
\hline
1  & \parbox[t]{0.70\linewidth}{\raggedright\ttfamily naver-clova-ix/donut-base} & 417 & 28.21 \\
2  & \parbox[t]{0.70\linewidth}{\raggedright\ttfamily llava-hf/llava-v1.6-mistral-7b-hf} & 167 & 11.30 \\
3  & \parbox[t]{0.70\linewidth}{\raggedright\ttfamily Qwen/Qwen2.5-VL-3B-Instruct} & 138 & 9.34 \\
4  & \parbox[t]{0.70\linewidth}{\raggedright\ttfamily Qwen/Qwen2.5-VL-7B-Instruct} & 127 & 8.59 \\
5  & \parbox[t]{0.70\linewidth}{\raggedright\ttfamily microsoft/git-base} & 107 & 7.24 \\
6  & \parbox[t]{0.70\linewidth}{\raggedright\ttfamily Qwen/Qwen2-VL-2B-Instruct} & 95 & 6.43 \\
7  & \parbox[t]{0.70\linewidth}{\raggedright\ttfamily Qwen/Qwen2-VL-7B-Instruct} & 70 & 4.74 \\
8  & \parbox[t]{0.70\linewidth}{\raggedright\ttfamily unsloth/gemma-3-4b-it-unsloth-bnb-4bit} & 52 & 3.52 \\
9  & \parbox[t]{0.70\linewidth}{\raggedright\ttfamily google/gemma-3-27b-it} & 52 & 3.52 \\
10 & \parbox[t]{0.70\linewidth}{\raggedright\ttfamily google/gemma-3-4b-it} & 51 & 3.45 \\
11 & \parbox[t]{0.70\linewidth}{\raggedright\ttfamily google/paligemma2-3b-pt-224} & 48 & 3.25 \\
12 & \parbox[t]{0.70\linewidth}{\raggedright\ttfamily google/paligemma-3b-pt-224} & 45 & 3.04 \\
13 & \parbox[t]{0.70\linewidth}{\raggedright\ttfamily OpenGVLab/InternVL3-1B-Instruct} & 42 & 2.84 \\
14 & \parbox[t]{0.70\linewidth}{\raggedright\ttfamily meta-llama/Llama-3.2-11B-Vision-Instruct} & 34 & 2.30 \\
15 & \parbox[t]{0.70\linewidth}{\raggedright\ttfamily google/gemma-3-12b-it} & 33 & 2.23 \\
\hline\hline
\end{tabular}

\vspace{0.20em} 
\footnotesize\emph{Note:} Edges are counted over recorded \texttt{finetune\_parent} relations where both parent and child are VLM-tagged.
\end{table*}

\section{Discussion}

Our results characterize a distinctive mode of multimodal innovation in the open LLM ecosystem: multimodality tends to enter major open families through a small number of VLM ``founders'' and then expands predominantly within VLM lineages, while direct lineage transitions from text-generation checkpoints to VLM tasks remain rare. This pattern is naturally described by founder effects and punctuated entry, concepts that have long been used to explain rapid change following rare founding events in biological evolution \cite{Mayr1954FounderEffect,Templeton1980FounderPrinciple} and, more broadly, punctuated evolutionary dynamics \cite{EldredgeGould1972Punctuated}. In the open model ecosystem, the ``founding'' event corresponds operationally to the appearance of new VLM roots without recorded parents and the subsequent concentration of derivative activity within those descendant lineages.

A key implication is that ecosystem-level availability of multimodal artifacts does not translate automatically into within-family diffusion. Cross-modal models and tasks are present in the broader hub earlier than they become prevalent inside major open LLM families, indicating a decoupling between global supply and family-level adoption. In diffusion terms, the relevant ``population'' is not the entire hub but the subset of lineages that define prominent families, and the bottleneck is the appearance of bridging mechanisms into those lineage trees \cite{Rogers2003Diffusion}. This is consistent with multimodality requiring additional inputs beyond those needed for text specialization, including multimodal data pipelines, architectural interfaces between vision encoders and language backbones, and evaluation tooling for grounding and visual faithfulness.

The rarity of text-generation$\rightarrow$VLM transitions along recorded parent--child edges should not be read as evidence that text-only innovations are irrelevant to multimodal systems. Many VLMs reuse a text backbone, so improvements in language pretraining, post-training, and efficiency can plausibly carry into VLM performance when integrated into a multimodal stack. Rather, our edge-level estimates indicate that this reuse rarely manifests as \emph{routine} lineage conversion via standard derivative operations. Fine-tuning, merging, and quantization are largely modality-preserving transformations, so the introduction of a vision channel typically requires a higher-complexity integration step that is not captured as incremental task mutation. This interpretation aligns with work on technological improvement showing that the rate and character of progress depend on underlying structural constraints and complexity \cite{McNerney2011DesignComplexity}. It also suggests that the emergence of widely adopted, standardized interfaces could change the observed regime by lowering integration costs---a hypothesis consistent with theories of modularity and architectural decomposition \cite{BaldwinClark2000DesignRules}.

These findings help explain why ``innovation diffusion'' in open LLM families can appear bursty even when underlying methods progress continuously. Founder-driven dynamics create path dependence: early successful VLM founders become disproportionately important conduits for downstream derivatives, concentrating subsequent innovation within a few lineages. This mechanism can accelerate within-lineage diffusion (rapid amplification through fine-tunes, quantization, and merges) while slowing cross-lineage diffusion (rare bridging events). Practically, this implies that improvements in text-only families will not necessarily propagate quickly into multimodal variants unless explicit integration work is performed to create new VLM-capable descendants or new VLM founders.

The results also suggest testable predictions about the future evolution of the open ecosystem. If the community develops more standardized, low-friction ways to attach and train vision modules---for example via parameter-efficient tuning (adapters; LoRA-family methods) and quantization-aware workflows---then the measured lineage transition rates from text-generation checkpoints to VLM tasks should rise \cite{Houlsby2019Adapters,Hu2021LoRA,Dettmers2023QLoRA}. Conversely, if multimodality continues to require bespoke pipelines and substantial engineering, then growth should remain dominated by within-VLM reproduction and periodic founder entry, and new multimodal capabilities should appear as bursts following the release of new VLM founders rather than as gradual conversion of many text-only branches.

Several limitations bound interpretation. First, lineage relations are self-reported metadata and are plausibly incomplete; missing parent annotations can inflate the apparent share of new ``roots'' and can undercount cross-type transfer that occurred but was not recorded. Second, task tags and model-card signals are noisy and heterogeneous, and our measurements should be interpreted as ecosystem-level indicators of intended use rather than ground-truth capability. Third, time-resolved analyses begin effectively in March 2022 due to timestamp backfilling, and family identification relies on name-based proxies, which can misclassify edge cases. Finally, our analysis characterizes diffusion of multimodality as recorded in metadata and lineages, not capability scaling as measured by standardized benchmarks. Integrating benchmark evidence and weight-level architectural parsing would strengthen causal attribution about which technical innovations transfer and how.

Overall, the picture that emerges is an ecosystem in which multimodality is shaped by founder effects: rare integration events establish VLM founders, and subsequent multimodal innovation diffuses mainly through within-lineage derivative activity. This mechanism provides a parsimonious explanation for delayed adoption within major open families despite earlier ecosystem-wide availability, and it yields concrete, measurable predictions about how improved modular tooling and reporting standards could change the evolutionary dynamics of multimodality in the open model ecosystem.

\section{Materials and Methods}
All analyses were conducted using the ModelBiome AI Ecosystem dataset (July 2025 snapshot), which aggregates public Hugging Face model metadata, task tags, model cards, and recorded parent--child lineage relations. The dataset comprises approximately $1.86\times10^{6}$ model entries and $3.02\times10^{6}$ directed lineage edges. Data processing and analysis were performed in Python using a reproducible Google Colab workflow. Models were assigned task categories using Hugging Face pipeline tags, with vision--language models (VLMs) defined as image$\leftrightarrow$text tasks (e.g., \texttt{image-to-text}, \texttt{image-text-to-text}). Open LLM families were identified via name-based proxies within Transformer architectures, excluding diffusion-oriented pipelines. Lineage-conditioned transition rates were computed over recorded relation types (fine-tuning, merging, adapters, quantization) and binned by child model creation month. Reported proportions use 95\% Wilson score confidence intervals. Code is available at \url{https://github.com/manuelcebrianramos/open-llm-multimodality-dynamics}.



\section*{Acknowledgments}
OpenAI ChatGPT was used for limited editorial and tooling support, including (i) sentence- and paragraph-level streamlining; (ii) generation of a subset of code; and (iii) identification of potential coding errors and refactoring suggestions. All interpretations and writing decisions were made by the author. Any LLM-generated material was reviewed and edited for accuracy and remains the author’s responsibility.




\bibliographystyle{apsrev4-2}
\bibliography{references}


\end{document}